# Smart Navigation System for Parking Assignment at Large Events: Incorporating Heterogeneous Driver Characteristics

Xi Cheng, Gaofeng Su, Siyuan Feng, Ke Liu, Chen Zhu, Hui Lin, Jilin Song, and Jianan Chen

*Abstract*—Parking challenges escalate significantly during large events such as concerts or sports games, yet few studies address dynamic parking lot assignments for such occasions. This paper introduces a smart navigation system designed to optimize parking assignments swiftly during large events, utilizing a mixed search algorithm that accounts for the heterogeneous characteristics of drivers. We conducted simulations in the Berkeley city area during the "Big Game" to validate our system and demonstrate the benefits of our innovative parking assignment approach.

*Index items - parking assignment, mixed search algorithm, big event*

## I. INTRODUCTION

With the rapid growth in urbanization, more and more people are streaming into big cities, which causes the heavier car congestion in the metropolitan area, leading to the shortage of parking. The shortage of parking space is considered as one of the major issues in the city transportation management because of the expensive construction cost of new parking lots under the circumstance of the limited spatial resource of a city. The thing becomes even worse when there is a big event going on or when citizens head to a central business district. According to one recent research work, Arnott et al. (2005) [1] mentioned that about 30% of cars on the road in the downtown area of major cities seemed to be cruising for parking spots, which took an average of 7.8 min. Another study (Soup, 2007) [2] found that the wandering of cars in order to find a parking facility is responsible for about 30% of the entire traffic in a city.

However, fortunately, due to the prevalence of smart devices and intelligent infrastructure, available parking spaces can be revealed and kept the drivers informed via an intelligent parking guidance system, which is capable of presenting paths to drivers in the selection of parking lots and direction guidance. This paper proposes a method to evaluate the benefit of a parking guidance algorithm to assign a specific parking lot for the drivers who drive to a destination where a big event happens. The parking guidance algorithm is proposed to assign the vehicles to the most appropriate parking lot considering distance to the destination and walk distance from the parking lots to the destination. In order to evaluate this optimal assignment method, a kind of method to simulate the parking in the real life is proposed and used which combines several searching potential and possible routes.

Parking assignment issues have been studied for a long time and many related works have been done by researchers. Some parking assignment methods as well as how to evaluate the impact of the implementation of these methods have been proposed. To evaluate the benefit of the parking assignment methods including some parking guidance information system, the choices of drivers have to be simulated. In order to model the parking choice considering behavior, a parking lot choice model was derived by using the logit function and the parameter calibration was conducted via a case study (Sattayhatewa and Smith, 2003) [3]. However, the paper did not use a method to evaluate how well the model worked. Parking guidance (Shin and Jun, 2014) [4], have been explored to help drivers find open parking spaces quickly. This study focuses on developing an advanced parking navigation system that can guide drivers to unutilized parking spaces and substantially reduce their cruising time for parking. And to evaluate the effectiveness of the proposed algorithm, simulation experiments have been carried out. In the simulation, they tested for the base case which assumes the situation that all drivers are heading to the nearest parking facility from their destinations without the use of our system. Guo et al. (2013) [5] gave out two types of parking choice models, a static game theoretic model and a dynamic neo-additive capacity model. Validation results showed higher predictive accuracy for the dynamic neo-additive capacity model compared to the static game theoretic model. However, this paper just used a small dataset to calibrate the parameters in the dynamic neo-additive model. Chen et al. (2019) [6] developed a novel parking navigation system for downtown parking that aims to guide drivers to their own most appropriate parking spaces (if any) and simulation experiments were conducted to demonstrate the capability of the proposed navigation system on reducing driving time. In the simulation section, matching system proposed by the authors was compared with status quo, greedy system as well as gravitational system. The gravitational approach was presented to guide drivers to parking spaces by Ayala et al. (2012a) [7]. The gravitational strategy described an incomplete context that drivers have no knowledge of the others. In this context, drivers make some prior probabilistic assumptions about the locations of the other vehicles in the game and the analysis is performed based on the expectations given by the prior distributions. Rehena et al. (2018) [8] developed a Multiple Criteria based Parking space Reservation algorithm taking user preferences into account

X. Cheng is with the University of Illinois at Chicago, Chicago, IL 60607, USA.
G. Su and K. Liu are with the University of California at Berkeley, Berkeley, CA 94702, USA.
S. Feng is with the Hong Kong University of Science and Technology, Clear Water Bay, Hong Kong.
C. Zhu (corresponding author) is with the Tsinghua University, Beijing, 100084, P. R. China.
H. Lin is with Northwestern University, Evanston, IL, 60208, USA.
J. Song is with the University of Toronto, Toronto, Ontario M5S 1A1 Canada.
J. Chen is with the University of British Columbia, Vancouver, BC, Canada V6T 1Z2.

and simulations for three set of user preferences represented by multiple criteria: distance between parking area and the destination, price per hour for reserving the space as well as unoccupied space for each parking area.

In this paper, heterogenous characteristics of drivers have been taken into consideration and thereby a mixed search algorithm was proposed. Based on these combination of different searching methods, each vehicle is assigned to a specific searching route. Simulations are done for this base case to evaluate the weight of choices of different drivers. However, all the studies mentioned above, and related works have not thought about assigning parking lots to vehicles combining several different searching routes together.

The structure of this paper is organized as follows. The details of the parking guidance algorithm to assign the vehicles to the most suitable parking lots are explained in Section 2. Section 3 conducts the method to evaluate the benefit of the proposed assignment algorithm using the baseline of several combined direction guidance ways (i.e. searching for parking starting from the nearest parking lots to the destination, starting from the nearest one to the current location, and starting from the nearby parking lots randomly). Results are displayed and discussed in section 4, which validate the feasibility and effectiveness of this evaluation method. Section 5 discusses and concludes this evaluation method.

## II. METHODOLOGY

### A. Parking assignment optimization method

In city areas where many cars are searching for parking spaces, it's crucial to efficiently assign each car to the most suitable parking lot. This can significantly reduce vehicle miles travelled (VMT), traffic caused by parking searches, and associated environmental issues. A parking assignment method is proposed here to minimize the time spent by drivers searching for parking.

This assignment method is an optimization problem that assigns each vehicle to the most suitable parking lot. However, it doesn't consider the travel time from the departure place to the assigned parking area, as the focus is on the parking aspect of the trips.

The goal of this optimization problem is to minimize the time spent from entry points to the parking lot, searching for a parking spot and the time walking from the parking lot to the destination. The first constraint ensures that each vehicle can be assigned to a specific parking lot. The second set of constraints are capacity constraints, which ensure that the number of vehicles assigned to parking lot i does not exceed its capacity. The final set includes binary constraints. This optimization problem is formulated as follows:

$$\min \sum_{i=1}^{m} \sum_{k=1}^{K} x_{ik}(TD_{ik} + TS_{ik} + TW_i)$$
$$s.t. \sum_{i=1}^{m} \sum_{k=1}^{K} x_{ik} = K$$
$$\sum_{k=1}^{K} x_{ik} \leq Vol\_Lot_i$$
$$x_{ik} = \begin{cases} 1 & if\ vehicle\ k\ is\ assigned\ to\ lot\ i \\ 0 & o.w. \end{cases}$$

Where :

$TD_{ik}$ is the driving time from the entry points, which is appointed to the vehicle $k$ randomly in advance, to the parking spot $i$ of the vehicle $k$.

$TS_{ik}$: search time in parking lot $i$ of vehicle $k$, which is the duration from entering the parking lot to exiting.

$TW_i$: walking time from parking lot $i$ to the destination.

Assume all drivers park in the space at parking lot $i$ in a sequence and the spaces are shown as Figure 1.

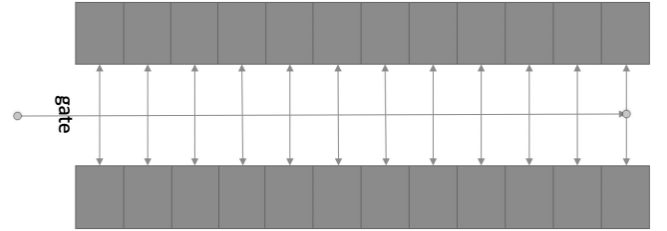

Figure 1. Layout of inner parking lots

$$TS_{ik} = (\frac{W/2 \cdot Vol_{Lot_i} \cdot O_i}{v_c} + \frac{W/2 \cdot Vol_{Lot_i} \cdot O_i}{v_w} + t_{stop}) + (N_i - 1) \cdot (\frac{R_{iL}}{v_{ud}} C_{iN} \cdot O_{iN} + t_{turn})$$

Where:

$O_i$: occupancy of parking lot $i$

$N_i$: number of floors of parking lot $i$

$O_{iN}$: occupancy of $N^{th}$ floor of parking lot $i$

$C_{iN}$: capacity of $N^{th}$ floor of parking lot $i$

$Vol\_Lot_i$: capacity of parking lot $i$

$R_{iL}$: length of a ramp of parking lot $i$

Parameters:

W: width of a spot

$v_c$: average cruise speed on the flat

$v_w$: average walk speed

$t_{stop}$: average parking and alighting time (assume 1 min)

$t_{turn}$: average U turn time (assume 10 s) between ramps

$v_{ud}$: average speed on and off ramp

### B. Parking Choices Reflecting Driver Heterogeneity

Parking choices in the real world typically arise from a complex interaction between individual drivers' parking preferences, their knowledge of the parking stock, the immediate availability of parking, and current traffic conditions [9]. This paper proposes a method for parking search, aiming to simulate real-life situations when drivers are unfamiliar with the parking options near their destination.

Past studies have simulated potential real-life situations to evaluate proposed parking assignment methods, assuming that all drivers follow a single method. These methods may include randomly choosing a parking lot near the destination, sequentially parking from the nearest lot to the destination, or selecting the nearest lot from their current location.

This study, however, blends different methods by assigning varying weights to each, intending to simulate real life more accurately. The realism of this assumed simulation needs to be validated in subsequent sections.

Suppose vehicles are categorized based on $l$ different search methods. Consider a rectangular region encompassing all parking lots near a destination as the study area, where Figure 2 is an example in the experimental setup. This region is accessible via m traffic links. Vehicles randomly enter the region through these m links, following an arrival time distribution. The arrival time of each vehicle is expressed as:

$$AT_i = ET_i + NS_i$$

Where,

$AT_i$: actual arrival time of vehicle *i*.

$ET_i$: expected arrival time of vehicle *i*, which conforms to poisson distribution.

$NS_i$: noise of arrival time of vehicle *i*, which conforms to normal distribution.

In our model, vehicles are categorized into *l* distinct search groups, with each vehicle randomly assigned to one of these groups. Each group employs a specific strategy for selecting parking lots. If a vehicle encounters a full parking lot, it is redirected to the next most suitable lot. However, a critical behavioral aspect is that drivers have limited tolerance for repeatedly encountering full lots. Many prefer to seek parking further away, where availability is higher but potentially less convenient.

To quantify drivers' tolerance, we model the likelihood of a driver abandoning their search for parking within a specific urban area as a function of time spent searching. This probability follows a gamma distribution, reflecting the variability in driver behavior across different scenarios and destinations. As a result, prolonged search times can lead to drivers exiting the designated parking area, exacerbating traffic congestion, increasing emissions, and wasting time.

This model allows us to simulate and analyze the impact of parking availability and search strategies on urban congestion and environmental outcomes. By understanding the limits of driver patience and the factors influencing their decisions, we can refine our parking assignment algorithms to minimize negative impacts and improve the efficiency of urban transportation networks.

III. EXPERIMENTAL SETUP

A significant influx of American football fans drives to the annual game between UC Berkeley and Stanford, held at California Memorial Stadium. Despite some fans' familiarity with Berkeley, the arrival of many new visitors complicates parking decisions for all. In such scenarios, even local drivers struggle to identify optimal parking spots due to changing conditions. To address this, our model treats all drivers as unfamiliar with the area to simplify guidance to suitable parking spaces, enhancing safety by reducing reliance on maps which can distract drivers.

Comprehensive data on the parking infrastructure around California Memorial Stadium has been collected. This includes the exact locations and capacities of all visitor parking lots. The dataset reveals that there are 21 parking lots within a designated rectangular region around the stadium, providing a total of 3,992 parking spaces (Figure 2).

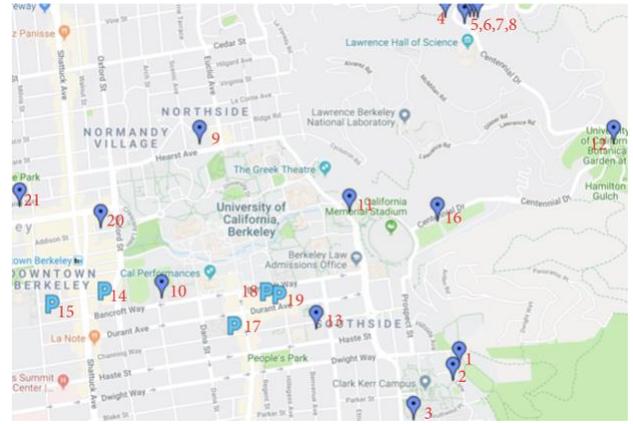

Figure 2. Case region around california memorial stadium.

We assume that the demand for parking exactly matches the supply, with all 3,992 parking spaces utilized. The parking assignment methodology described earlier assigns each vehicle to a specific parking space, optimizing the total time spent from entry points to parked position. This system assumes all vehicles arrive uniformly, beginning two hours before the event via 12 access streets.

To assess the effectiveness of our parking assignment system, we simulate realistic parking scenarios. This involves assuming that all vehicles attempt to enter the parking region simultaneously, mimicking the conditions on a game day. By comparing this scenario with our optimized assignment system, we evaluate the potential time savings and reduced congestion offered by our approach. We maintain consistency in our simulation by matching the number of vehicles to the total number of available parking spaces.

To effectively manage parking during the event, vehicles are divided into four search groups, each employing a distinct strategy for selecting parking lots., which have the same weight.

- Group 1: This group aims to find the nearest available parking lot as soon as the vehicle enters the designated region. This strategy prioritizes minimizing the driving distance from the region's entry points to the parking lots.
- Group 2: This group of vehicles search for the nearest parking lot upon entering the region. However, this group excludes lots located close to the stadium to avoid congestion in high-demand areas and spread-out parking utilization.
- Group 3: Vehicles in this group search for the nearest parking lot relative to the stadium itself. This approach focuses on reducing the walking distance from the parking lot to the event, prioritizing convenience for attendees.
- Group 4: This comprehensive strategy considers both the driving distance to the parking lot from the current vehicle location and the subsequent walking distance from the parking lot to the stadium. The total travel time (driving plus walking) is minimized, providing an optimized parking solution based on overall accessibility.

To validate the effectiveness of our base case scenario, we conducted comparative analyses with foundational models established in previous studies. Specifically, Shin and Jun [4] employed a model where all drivers aim for the nearest parking facility from their destination without any guidance system—mirroring the strategy used by Group 3 in our study. Conversely, Rehena [8] designed a base case assuming that drivers choose parking lots based on the total travel time, which includes both the drive to the parking lot and the subsequent walk to the destination, aligning with the approach of Group 4 in our research.

For accurate evaluation and simulation, the distances between parking lots within a designated rectangular region around California Memorial Stadium are crucial, as illustrated in Figure 2. To facilitate the computation of these distances and the subsequent travel times for each vehicle, we converted geographic coordinates into Cartesian coordinates using Miller projections [11]. This conversion simplifies the distance calculations between any given parking lot and the stadium, enhancing the precision of our simulation results.

$$x' = M_x(lon) = lon$$
$$y' = M_y(lat) = 1.25 \cdot ln(tan(\pi/4 + 0.4 \cdot lat))$$

According to the Miller projections, I apply the following transformation to get from *(lon, lat)* to *(x, y)*.

$$L = 6381372 \cdot \pi \cdot 2$$
$$x = L/2 + L/(2 \cdot \pi) \cdot x'$$
$$y = L/4 - L/(4 \cdot mill) \cdot y'$$

where:
mill is a constant which is set as 2.3.

With the help of the easy-to-calculate Cartesian coordinate system, the distance from every parking lot to the stadium is calculated. Reasonably and applicably, assume the sideways taken into consider are vertical with each other (see Figure 2).

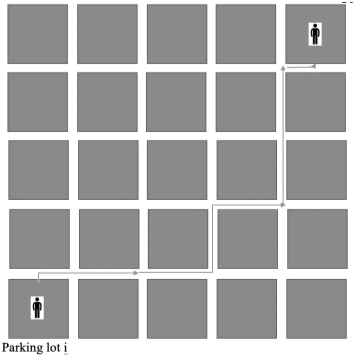

Figure 3. Distance calculation demonstration.

Therefore, the Manhattan distance between parking lot i and parking lot j or the destination is going to be equal to

$$\|x_i - x_j\| + |y_i - y_j|$$

where:
$x_i, x_j; y_i, y_j$ are the horizontal and vertical coordinates of the parking lot *i* and parking lot *j*.

In this way, the distance matrix for the use of problem developing was generated.

Since we don't know the arrival time of each user, we need to define a set of possible arrival times. The optimization problem is framed for a rush hour period from 10 a.m. to 12 p.m. To simplify the problem, we divide this total time into 12 segments using 10-minute intervals.

While the arrival time for each individual can be arbitrarily assigned, it's more realistic to simulate a scenario where most people arrive around 11 a.m. To do this, we use a Poisson distribution model to simulate the expected arrival time for each individual, adding normally distributed noise to each timestamp.

We generate a number based on the density function of the Poisson distribution, which represents the arrival time. However, one drawback of the Poisson model is that the generated number isn't bounded—it could theoretically go to infinity. To manage this, we limit the range of the number and rescale it to fit within our 10 a.m. to 12 p.m. time frame. The final arrangement of vehicle arrival times, including the added noise, is illustrated in Figure 4.

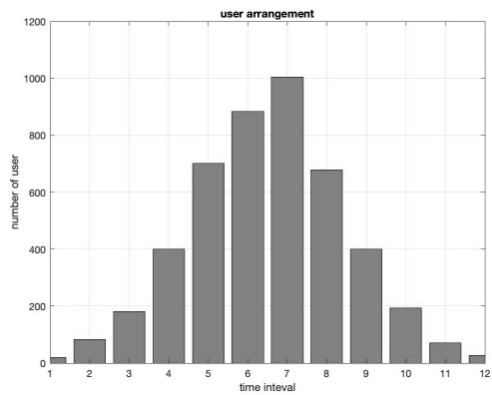

Figure 4. Arrival vehicles arrangement.

IV. RESULTS

Utilizing the arrival times and parking strategies detailed earlier, we conducted simulations to model the parking process within the designated case region. Our analysis focused on comparing different group strategies against the optimal assignment results. From these comparisons, we found that the average rerouting time wasted per vehicle is 4.7 minutes. This metric serves as a critical indicator of the efficiency of our parking management system.

Due to the inherent uncertainty and randomness in vehicle arrival times, multiple simulations were necessary to ensure the robustness of our findings. These simulations helped determine the required number of iterations to achieve reliable results. We analyzed how the average rerouting time varied across different simulation runs to understand the stability and consistency of our model. The relationship between the average rerouting time and the number of simulations conducted is depicted in Figure 5.

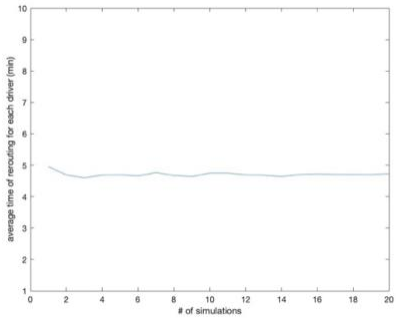

Figure 5.  Relation of average rerouting time and # of simulations.

To demonstrate the validity and effectiveness of this combination method, a comparison is essential. Shin's method (i.e., Group 3) and Rehena's base case method (i.e., Group 4) are also simulated, and the comparison is presented in Table 1.

TABLE I.  COMPARISON OF BASE CASES OF AVERAGE REROUTED TIME FOR EACH DRIVER (MIN)

| Comparisons | Shin's method | Rehena's method | This study |
|---|---|---|---|
| Time (min) | 14.6 | 6.1 | 4.7 |

Table 1 indicates that the base case of this study demonstrates a significant improvement in the average rerouted time, implying considerable time savings for parking. However, due to the difficulty or even impossibility of simulating real parking behavior, there are no detailed standards to judge these benchmarks and determine which is better for evaluating the benefits of the proposed assignment method.

Comparing the number of failed searches is a more valid criterion for contrasting these different methods. Some drivers fail to park in this region and detour to further areas, which can be defined as a failed search. The correlation between the number of failed searches in different parking lots and the number of simulations is also displayed in Figure 6. After around 10 times of simulations, the average rerouting time and the number of failed searches tend to be stable, which means 10 times of simulation is enough for simulating this realistic baseline.

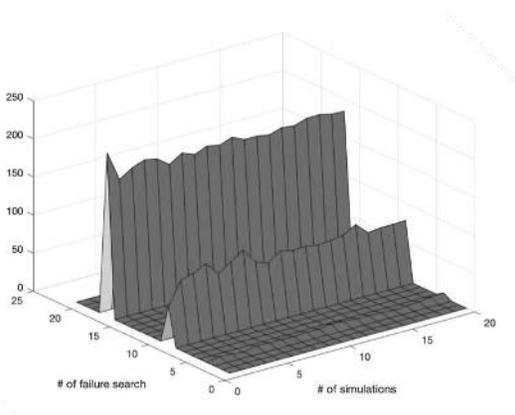

Figure 6. Failure search with different parking lot and # simulations.

According to the figure 7, there are around 150 and 50 unutilized spots in parking lot 8 and 12 respectively. Other parking lots are completely utilized after the simulation. However, this can be explained that these two parking lots are the farthest to the memorial stadium compared to the other parking lots. It cannot be acceptable that there are plenty of spots remaining in two parking lots while other parking lots are fully utilized.

The tolerance for each driver to spend time on parking is different. Some of them leave this region to another place to park with available spots after only searching for two or three full parking structures. Some of them can tolerate searching for 7 or 8 filled parking lots. Others can even tolerate searching for more than 10 filled lots. Afterwards, those drivers who is unable to find a parking lot leave this parking region, which accounts for a failed search. Correspondingly, there is supposed to be an unutilized parking spot in the end. Therefore, some of the parking lots cannot filled up in the end.

Then, to make the simulation much more reasonable, 20% of the vehicles are assigned to the parking lots in the region randomly. After certain times of simulation, the relation between failed search in different parking lot and the times of simulation is shown in Figure 7.

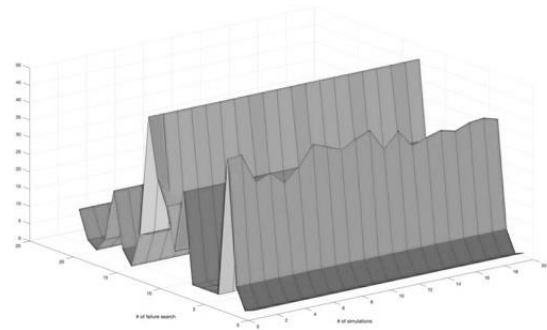

Figure 7. Failed searching with different parking lot and times of simulations (20% randomly).

From Figure 7, several parking lots are filled up with vehicles, while others have some spaces left compared to the capacities after the assignment, which is reasonably utilized to some extent. For example, parking lot 13 has the most spaces. Since this combined method takes a portion of randomness into account, there are the most failed searches in parking lot 13.

With the change of number of times of simulation, the number of failed searches fluctuates a little bit. Therefore, the failed searches result of only one time of simulation is able to be the representative of 20 times of simulation.

Similarly, results of the number of failed searches of Shin's method and Rehena's method are also shown as Figure 8.

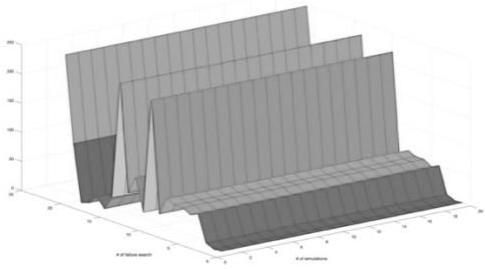

Figure 8a: Shin's method

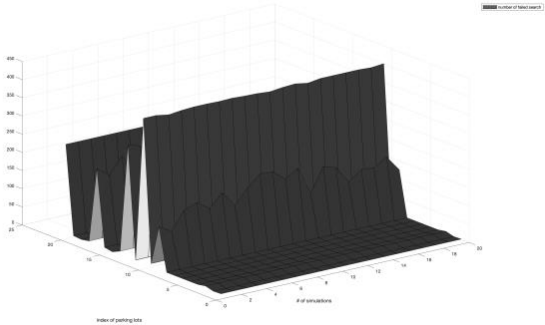

Figure 8b: Rehena's method

Figure 8. Compare to other methods: 8a: Shin's method & 8b: Rehena's method

According to the Figure 8a, there are over 200 parking spaces remaining in the parking lot 10, 14 and 21. Shin's method brings about much more failed searches compared to the proposed methods. Plus, average rerouted time of each driver of Shin's method is 14.6 minutes, which is more than 3 times of the method proposed by this study.

From the Figure 8b, there are over 200 parking spaces remaining in the parking lot 11, 13 and 21. Especially in the parking lot 11, the unutilized parking spaces are round 400. Rehena's method is not that realistic, since some parking lots cannot have so many parking

In conclusion, if we use Shin's method or Rehena's method without combination, several parking lots cannot be utilized since they are located in the back of those two methods' order. When the time for parking is in excess of the tolerance threshold, vehicles would give up searching for parking, which brings about a large number of failed searches. Therefore, the proposed combined base method is much more realistic and effective to evaluate the benefit of an optimization methodology.

## V. CONCLUSION

This paper introduces a refined parking assignment method that strategically allocates vehicles to designated parking structures, coupled with a comprehensive simulation of real-life parking scenarios. By considering the heterogeneous characteristics of drivers, our optimization method significantly reduces rerouting times, offering a robust improvement over traditional baselines that mirror actual parking behaviors.

Unlike many previous studies that relied on theoretical models with unvalidated parameters, our approach leverages actual parking structure data and accounts for the varied preferences and behaviors of drivers to more accurately estimate search times within parking facilities. This methodological innovation allows for a more realistic assessment of parking management strategies.

Furthermore, we have developed a simulated baseline that closely approximates real-life conditions, providing a strong foundation for evaluating the efficacy of our parking assignment optimization. This is a significant advancement, as many prior studies did not include detailed quantitative comparisons between optimized methods and realistic baselines, often presuming the superiority of any assignment method over unguided vehicle parking.

However, it is crucial to acknowledge that while our simulation represents a significant step forward, it does not capture all aspects of real-life driver behavior with complete accuracy. The complexity of actual parking dynamics poses substantial challenges to any simulation model. Future research should therefore focus on refining these simulations to better reflect true driver behaviors and parking patterns, possibly integrating emerging technologies and data analytics to enhance precision and applicability. There remains ample scope for further exploration in this field, particularly in terms of improving the realism of simulation models and exploring new variables that influence parking behavior. Continued advancements in this area will not only refine our understanding of effective parking management but also contribute to the broader field of urban traffic optimization.

REFERENCES

[1] G. O. Young, "Synthetic structure of industrial plastics (Book style with paper title and editor)," in *Plastics*, 2nd ed. vol. 3, J. Peters, Ed. New York: McGraw-Hill, 1964, pp. 15–64.
[2] Soup, D., 2007. Cruising for parking. Access 30, 16–22.
[3] Sattayhatewa, P. and Smith, R. Development of Parking Choice Models for Special Events. Transportation Research Record 1858, Paper No. 03-2295.
[4] Shin, J. and Jun, H. A study on smart parking guidance algorithm. Transportation Research Part C 44 (2014) 299–317.
[5] Guo, L., Huang, S., Zhuang, J., Sadek, A., 2012. Modeling Parking Behavior Under Uncertainty: A Static Game Theoretic versus a Sequential Neo-additive Capacity Modeling Approach. Networks and Spatial Economics September 2013, Volume 13, Issue 3, pp 327–350.
[6] Chen, Z., Spana, S., Yin, Y., Du, Y., 2019. Networks and Spatial Economics.
[7] Ayala D, Wolfson O, Xu B, DasGupta B, Lin J (2012a) Parking in competitive settings: a gravitational
[8] approach. In Mobile Data Management (MDM), 2012 IEEE 13th International Conference on (pp 27–32). IEEE
[9] Rehena, Z., Mondal, A., Janssen, M., 2018. A Multiple-Criteria Algorithm for Smart Parking: Making fair and preferred parking reservations in Smart Cities.
[10] Polak, J. and Axhause, K., 1990. Parking Search Behaviour: A Review of Current Research and Future Prospects. Transport Studies Unit Oxford University.
[11] Snyder, J. P. (1993). Flattening the Earth. Two Thousand Years of Map Projections. Chicago and London: University of Chicago Press.